\documentclass[letterpaper, 10 pt, conference]{ieeeconf}  

\IEEEoverridecommandlockouts                              

\usepackage{graphics} 
\usepackage{epsfig} 
\usepackage{mathptmx} 
\usepackage{times} 
\usepackage{amsmath} 
\usepackage{amssymb}  
\usepackage{cite}
\usepackage{float}
\usepackage[linesnumbered,ruled]{algorithm2e}
\usepackage{subcaption}
\usepackage{stfloats}
\usepackage{color}
\newcommand{\n}[1]{{\color{black}{#1}}}
\newcommand{\m}[1]{{\color{black}{#1}}}
\usepackage{booktabs}
\usepackage[font=footnotesize]{caption} 
\usepackage{url}

\title{\LARGE \bf
UTTG: A Universal Teleoperation Framework via Online Trajectory Generation}

\author{Shengjian Fang\textsuperscript{1,\#}, Yixuan Zhou\textsuperscript{1,\#}, Yu Zheng\textsuperscript{1}, Pengyu Jiang\textsuperscript{1}, Siyuan Liu\textsuperscript{1}, and Hesheng Wang\textsuperscript{1,*}
\thanks{*This work was supported in part by the Natural Science Foundation of China under Grant 62225309, U24A20278, 62361166632, U21A20480 and 62403311. (Corresponding Author: Hesheng Wang)}
\thanks{$^{1}$ Authors are with the IRMV Lab, Department of Automation, Shanghai Jiao Tong University, Shanghai 200240, China. Emails: {fang20021005, yixuanzhou, zhengyu0730, jiangpengyu, siyuan\_l, wanghesheng}@sjtu.edu.cn}%
\thanks{\#These authors contributed to the work equallly and should be regarded as co-first authors.}}

\begin{document}

\maketitle
\thispagestyle{empty}
\pagestyle{empty}

\begin{abstract}
Teleoperation is crucial for hazardous environment operations and serves as a key tool for collecting expert demonstrations in robot learning. However, existing methods face robotic hardware dependency and control frequency mismatches between teleoperation devices and robotic platforms. Our approach automatically extracts kinematic parameters from unified robot description format (URDF) files, and enables pluggable deployment across diverse robots through uniform interfaces. The proposed interpolation algorithm bridges the frequency gap between low-rate human inputs and high-frequency robotic control commands through online continuous trajectory generation, \n{while requiring no access to the closed, bottom-level control loop}. To enhance trajectory smoothness, we introduce a minimum-stretch spline that optimizes the motion quality. The system further provides precision and rapid modes to accommodate different task requirements. Experiments across various robotic platforms including dual-arm ones demonstrate generality and smooth operation performance of our methods. The code is developed in C++ with python interface, and available at https://github.com/IRMV-Manipulation-Group/UTTG.
\end{abstract}

\section{INTRODUCTION}
Teleoperated robotic systems have become indispensable in hazardous environments, including construction sites and radioactive contamination zones, as these systems enhance human operational capacity while ensuring worker safety \cite{darvishTeleoperationHumanoidRobots2023}. Through remote task execution capabilities, such systems effectively mitigate occupational hazards and enhance operational efficiency. Furthermore, the recent advancement of data-driven approaches in robotics has amplified the importance of teleoperation, particularly for imitation learning (IL) frameworks \cite{ravichandarRecentAdvancesRobot2020}. IL enables robots to acquire dexterous manipulation skills through large-scale human demonstration datasets \cite{sicilianoSpringerHandbookRobotics2016}, where teleoperation systems serve as critical tools for curating expert level demonstrations and transmitting human knowledge to robotic platforms \cite{jangBCZZeroShotTask2022a}.

Recent advancements in teleoperation have primarily focused on two distinct research directions. One develops modular accessory kits \cite{wuGELLOGeneralLowCost2024} to augment commercial robotic platforms, yet such solutions frequently entail intricate fabrication procedures that hinder rapid deployment. The other employs learning-based agents that map human input trajectories to robot motion commands \cite{honerkamp2024zero}. However, training these agents necessitates constructing URDF-based simulation environments for each target platform, requiring parameter tuning and collision modeling that inevitably binds the agents to specific hardware configurations. These limitations restrict the general applicability and scalability of current methods as robotic platforms diversify.
\begin{figure}[t]
    \centering
    \includegraphics[width=\linewidth]{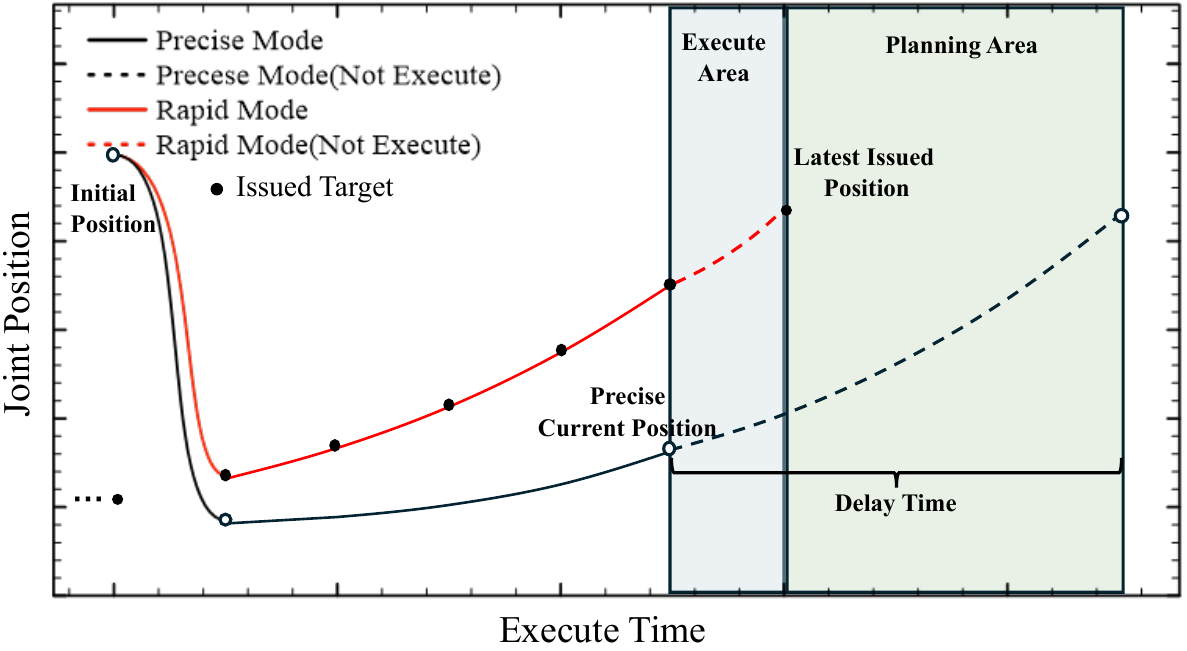}
    \caption{Illustrations of rapid mode and precise mode.}
    \label{two_modes}
\vspace{-5mm}
\end{figure}

To resolve these challenges, we propose a teleoperation framework featuring two capabilities. First, our framework resolves the control frequency mismatch between low-frequency human-input interfaces (e.g. 10-50Hz vision-based skeleton recognition) and high-frequency robotic platforms like Franka (typically 200-1000Hz) thereby supporting cross-platform deployment. Second, the method directly extracts critical kinematic parameters from URDF models and enables teleoperation through standardized I/O(input/output) interfaces and initial robot state, achieving rapid deployment and pluggable compatibility across heterogeneous robotic platforms. The key contributions of our framework are as follows:

\begin{enumerate}
\item An online trajectory generation method with cubic spline interpolation is proposed to reconcile heterogeneous frequency interfaces.
\item Standardized I/O interfaces with \n{automatic} URDF configuration \n{tool} enable plug-and-play teleoperation across robotic platforms. 
\item Two configurable operation modes (Fig. \ref{two_modes}), one optimized for precision and the other for speed, are provided. Experimental results in real world scenarios validate the effectiveness of our approach.
\end{enumerate}

\section{RELATED WORK}

\subsection{Teleoperation}

The teleoperation task involves converting human motion to robot joint commands, introducing significant dependencies on specific devices and robot models. Several control strategies have been proposed to address this issue. \textbf{Joint replication} directly maps operator movement to robot joint angles, establishing a one-to-one correspondence between human motion and robot actuation \cite{elsner2022parti, fuMobileALOHALearning2024}. Two main approaches exist: a master-slave structure and a robot-like exoskeleton design. Both require the teleoperation device and the robot to share similar structures, as the mapping depends on kinematic congruence. \textbf{Motion retargeting} adjusts operator input to accommodate robot limitations, such as joint constraints and workspace boundaries. This method typically uses exoskeletons resembling human structure \cite{ishiguro2020bilateral} or motion capture systems \cite{arduengo2021human, krebs2021kit} to capture human motion data. \textbf{End effector control} and \textbf{vision-based control} \cite{ding2024bunny} focus on controlling the position and orientation of the human hand. Unlike joint replication and motion retargeting, these methods do not rely on complex external devices. However, a major challenge for these approaches is that current robot hardware execution layers do not support low frequency inputs. Joint replication methods allow high frequency sampling, whereas motion retargeting and end effector control tend to operate at lower frequencies and often struggle to maintain continuous control.

To overcome these challenges, we develop a simplified and generalized teleoperation framework. The framework maintains the essential continuity and precision required for effective teleoperation, while reducing reliance on particular device and robot configurations.
\begin{figure*}[ht]
    \centering
    \includegraphics[width=0.8\textwidth]{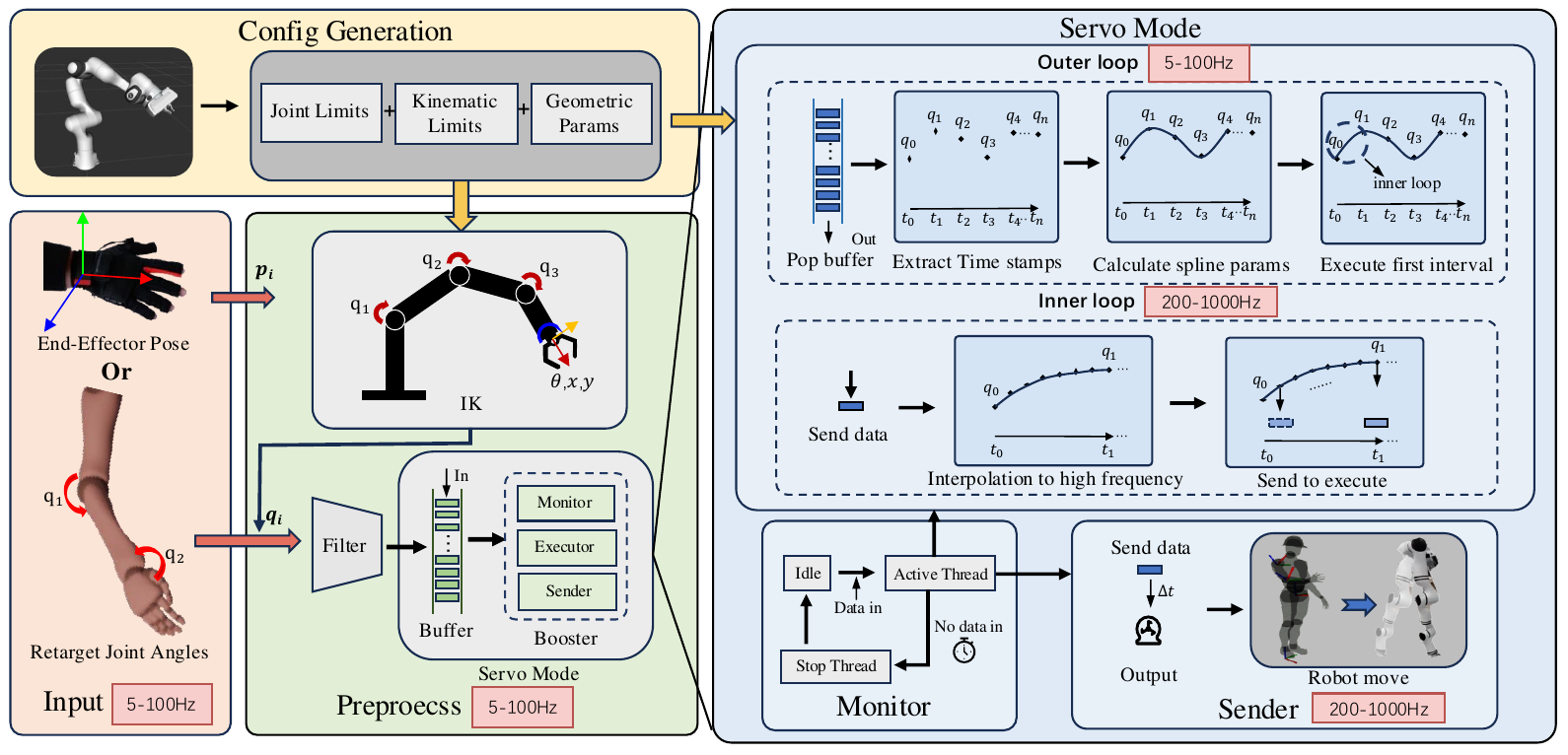} 
    \caption{UTTG framework: We utilize URDF of robot to automatically generate the required parameter files, decoupling human input from specific robot configurations. Our input can be either the end-effector pose or the joint angles. Using the generated files, we perform IK to compute the corresponding joint angles, and the output consists of high-frequency joint position commands for the robot.}
    \label{fig1}
    \vspace{-15pt}
\end{figure*}
\vspace{-2mm}
\subsection{Online Trajectory Generation}
A direct method for teleoperation is to design a PD (Proportional-Differential) controller requiring access to the bottom-level actuators. However, it is typically disabled by manufactures and provided as a specific hardware-dependent feature. Fortunately, the position-level real-time servo interface is a standard capability for modern robotic platforms. Thus, with the help of online trajectory generation, we can achieve teleoperation system evolving time-varying targets.

Traditional trajectory generation methods \cite{lavalle2006planning, pham2014general} employ offline computation paradigms that precompute entire trajectories before execution. These approaches fundamentally lack the capacity to handle real-time target updates or inherent system state disturbances in teleoperation scenarios. While existing online trajectory generation solutions partially address this limitation, analytical approaches \cite{broquere2008soft} often impose restrictive boundary conditions, and functional implementations \cite{berscheid2021jerk} are often commercial and complex. Consequently, we propose an open-source trajectory generation method that incrementally constructs motion paths from the current state to each new target point, continuously updating trajectories to satisfy teleoperation requirements.
\vspace{-1mm}
\section{METHOD}
 As illustrated in Fig. \ref{fig1}, the framework adopts a modular architecture comprising four components. The \textbf{Config Generation} module \n{automatically} extracts essential configuration information required for teleoperation from URDF. The \textbf{Input} module provides standardized interfaces for two input types: joint angles and end-effector poses, the latter being converted to joint space via an integrated IK solver. Both input streams \n{will} undergo filtering to remove unintended human tremors or noise in the \textbf{Preprocess} module. The filtered joint targets and their synchronized timestamps are stored in a buffer queue for subsequent trajectory planning. The \textbf{Servo Mode} module performs trajectory generation and execution by retrieving points from the buffer and generating smooth trajectories in precise or rapid modes based on the task requirements. The generated trajectories are dispatched to robotic actuators through standardized \n{position-level} interfaces for execution. To maintain stable and continuous trajectory execution, three threads operate in parallel: a monitor thread monitors system status, an executor thread generates trajectories, and a sender thread sends the resulting commands to the motors.


\subsection{Trajectory Generation}
\label{tg}
To interpolate discrete points, cubic spline is commonly adopted for trajectory interpolation due to their inherent smoothness and continuity. And its low computational cost is also suitable for online trajectory generation. 

As shown in \cite{TrajectoryPlanningAutomatic2009}, by connecting $n+1$ samples with $n$ segments of cubic splines and assigning each interval as $T_i$ for $i \in [0, n-1]$, the expression of the cubic spline on the $i$-th segment can be written as
\begin{equation}
\n{
    \mathbf{s}_i(t) = \mathbf{a}_i(\mathbf{m}_i) + \mathbf{b}_i(\mathbf{m}_i) t + \mathbf{c}_i(\mathbf{m}_i) t^2 + \mathbf{d}_i(\mathbf{m}_i) t^3,
    }
\end{equation}
where $t \in [0, T_i]$. Following \cite{zhouSpatiotemporalOptimalTrajectory2024}, the cubic spline problem can be solved by computing the \n{supplementary variables $\mathbf{m}_i$}. 

In the specific case examined by \cite{TrajectoryPlanningAutomatic2009}, where initial and final velocities and accelerations are assigned. In this case, two assistant points must be added, and time instants becomes
$$
\begin{aligned}
\mathbf{T} = [\overline{T}_0, \overline{T}_1,T_{1},\cdots,T_{n-2},\overline{T}_{n-2},\overline{T}_{n-1}],
\end{aligned}
$$
where $\overline{T}_0 =\beta T_0 , \overline{T}_1 = (1-\beta)T_0, \overline{T}_{n-1} =\beta T_{n-1} , \overline{T}_{n-2} = (1-\beta)T_{n-1}$. The band matrix $\mathbf{A}$ and coefficient matrix $\mathbf{C}$ are given by:
\begin{equation}
\resizebox{\columnwidth}{!}{$
    \mathbf{A} =
    \frac{1}{6}\begin{bmatrix}
    \frac{2-\beta}{1-\beta}T_0&(1-\beta) T_0 &&& \\
    \frac{1-2\beta}{1-\beta}T_0&2((1-\beta)T_0+T_1)&T_1 && \\
    &\ddots &\ddots &\ddots&&\\
    &&&(1-\beta)T_{n-1}&\frac{2-\beta}{1-\beta}{T}_{n-1}
    \end{bmatrix}
$}
\end{equation}
\begin{equation}
\resizebox{\columnwidth}{!}{$
\mathbf{C}= \begin{bmatrix}
-\frac{1}{1-\beta}T_0^{-1}&\frac{1}{1-\beta}T_0^{-1}\\
\frac{1}{1-\beta}T_0^{-1}&-\frac{1}{1-\beta}T_0^{-1}-T_1^{-1}&T_1^{-1}&&&\\
&\ddots&\ddots&\ddots&&\\
&&T_{n-2}^{-1}&-T_{n-2}^{-1}-\frac{1}{1-\beta}T_{n-1}^{-1}&\frac{1}{1-\beta}T_{n-1}^{-1}\\
&&&\frac{1}{1-\beta}T_{n-1}^{-1}&-\frac{1}{1-\beta}T_{n-1}^{-1}
\end{bmatrix}.
$}
\end{equation}
Then the second-derivative coefficient vector $\mathbf{m}$, obtained by solving the banded linear system, 
$
    \mathbf{m} = \mathbf{A}^{-1}(\mathbf{C}\mathbf{S}-\mathbf{D}),
$
uniquely determines the analytic expressions of all cubic spline segments through backward substitution of boundary conditions \n{under boundary condition $\mathbf{D}$}. The standard cubic splines excel at interpolating issued waypoints \n{$\mathbf{s}_i=\mathbf{q}_i$, where $\mathbf{s}_i$ represents the actual interpolated point, while $\mathbf{q}_i$ expresses the waypoint issued.} 

Practical teleoperation requires an intentional trade-off between path accuracy and motion quality. We introduce the min stretch cubic spline with assigned initial and end status, formulated to minimize:
\begin{equation}
\label{optimization}
\small{
    \textbf{S}^* = \n{\mathop{\operatorname{argmin}}\limits_{\mathbf{s}_i}}\left(\mu \sum_{i=0}^{n}w_i(\textbf{s}_i-\textbf{q}_i)^2 + (1-\mu)\sum_{i=0}^{n-1} \int_0^{T_i} \|\ddot{\mathbf{p}}_i(t)\|^2 dt\right)
    },
\end{equation}
where $\mu \in [0, 1]$ controls the importance between positional accuracy and smoothness, $w_i$ modify the weight of the $i$-th fitting error. The second term of  \n{\eqref{optimization}} can be simplified as $ \operatorname{tr}\left(\mathbf{m}^T \overline{\mathbf{A}} \mathbf{m}\right)$, and $\overline{\mathbf{A}}$ is \n{formulated} as:
\begin{equation}
\overline{\mathbf{A}} = \frac{1}{6}
\begin{bmatrix}
2T_0 & T_0 & & & \\
T_0 & 2(T_0 + T_1) & T_1 & & \\
& T_1 & 2(T_1 + T_2) & T_2 & \\
& & \ddots & \ddots & \ddots \\
& & & T_{n-1} & 2T_{n-1}
\end{bmatrix},
\end{equation}
where \(\operatorname{tr}(\cdot)\) denotes the trace of matrix. Then \eqref{optimization} \n{can be reformulated into a compact form as} 
\begin{equation}
\label{min-prob}
\begin{aligned}
\mathbf{S}^* = \n{\mathop{\operatorname{argmin}}\limits_{\mathbf{s}_i}}&\left( \mu\operatorname{tr}((\mathbf{Q}-\mathbf{S})^T\mathbf{W}(\mathbf{Q}-\mathbf{S})) + (1-\mu)\operatorname{tr}\left(\mathbf{m}^T \overline{\mathbf{A}} \mathbf{m}\right)\right)\\
&s.t. \mathbf{m} = \mathbf{A}^{-1} \left(\mathbf{C}\mathbf{S}-\mathbf{D}\right),
\end{aligned}
\end{equation}
where $\mathbf{W}$ \n{is the compact matrix form of $w_i$} and $\mathbf{S}$ represents the optimized trajectory waypoints that balance these objectives. \m{It should be noted that the current formulation focuses on achieving deterministic computation cycles, prioritizing real-time guarantees over Kinematic inequality constraints typically requiring iterative resolution \cite{MoveItDocs2023}.} 

\n{The problem \eqref{min-prob} can be solved effeciently with its convex property and modern QP-solver like piqp \cite{schwan2023piqp}. Significantly, when requiring motion between static configurations (with zero initial/final velocities and accelerations), the spline generation permits a simplified closed-form solution through boundary condition analysis, as we shall derive in subsequent sections.} 

Taking $\mathbf{D}$ $\equiv \mathbf{0}$ yields (which denotes zero initial/final velocities and accelerations) and substituting the constraint $\mathbf{m} = \mathbf{A}^{-1}\mathbf{C}\mathbf{S}$ into the objective function, we solve the resulting equality-constrained quadratic program by enforcing first-order optimality conditions:
\begin{equation}
\label{s_star}
    \mathbf{S^*} = \mathbf{Q}-\lambda \mathbf{W}^{-1}\mathbf{C}^T(\mathbf{G}^{-1}+\lambda \mathbf{C}\mathbf{W}^{-1}\mathbf{C}^T)^{-1}\mathbf{C}\mathbf{Q},
\end{equation}
with \(\lambda = \frac{1-\mu}{\mu}\) and \(\mathbf{G} = \mathbf{A}^{-1}\overline{\mathbf{A}}\mathbf{A}^{-1}\). The direct solution for $\mathbf{S^*}$ involves computationally expensive matrix inversions, prompting us to reformulate the optimization in terms of $\mathbf{m}$ through algebraic manipulation.
Suppose $(\mathbf{G}^{-1}+\lambda \mathbf{C}\mathbf{W}^{-1}\mathbf{C}^T)\mathbf{g} = \mathbf{C}\mathbf{Q}$, we substitute $\mathbf{S^*}$ from \eqref{s_star} into the constraint $\mathbf{m} = \mathbf{A}^{-1}\mathbf{C}\mathbf{S}$ and derive the simplified expression:
\begin{equation}
    \mathbf{m} = \mathbf{A}^{-1}(\mathbf{C}\mathbf{Q}-\lambda \mathbf{C}\mathbf{W}^{-1}\mathbf{C}^T\mathbf{g})=\overline{\mathbf{A}}^{-1}\mathbf{A}\mathbf{g}.
\end{equation}
For efficient computation, we suppose $\overline{\mathbf{A}}=\mathbf{A}$, $\mathbf{W}=\mathbf{I}$, and the system reduces to:
\begin{equation}
    \mathbf{m}=\mathbf{g}=(\mathbf{A}+\lambda \mathbf{C}\mathbf{C}^T)^{-1}\mathbf{C}\mathbf{Q},
\end{equation}
where $\mathbf{A}+\lambda \mathbf{C}\mathbf{C}^T$ is a band matrix. This sparsity structure enables efficient numerical methods for matrix inversion, improving computational speed and scalability of the proposed approach. 

Additionally, the proposed methodology intentionally excludes collision constraints, instead leveraging situational awareness of human operators for real-time obstacle avoidance through the teleoperation interface. 

\subsection{Framework for Precise Manipulation Task}
This subsection describes \n{our} framework for high precision manipulation tasks like fine assembly, surgical operation, and precision grasping. In such applications, the trajectory generated by the system must pass through each issued waypoint. The optimization weight $\mu$ in \eqref{min-prob} is consequently set close to unity ($\mu = 0.999$) to prioritize positional accuracy over trajectory smoothness. 
\begin{algorithm}[t]
\small
\caption{Precise Mode Trajectory Planning}
\label{precise}
\let\oldnl\nl
\newcommand{\nonl}{\renewcommand{\nl}{\let\nl\oldnl}}
\KwIn{$buffer$, $\mathcal{C}_{\text{robot}}$, $\Delta t_{\text{output}}$} 
\KwOut{$q_{\text{SendServo}}$} 

\textbf{Initialization:} $StartServo \gets \text{True}$, $FirstTime \gets \text{True}$

\BlankLine
\While{$StartServo = \text{True or } buffer \text{ is not empty}$}{
    $q_{\text{current}} \gets \text{GetCurrentPositions}$\;
    \eIf{$FirstTime$}{
        $q_{\text{first}} \gets \text{pop buffer}$\;
        $\sigma(\Xi, \tau) \gets \text{SolvePTP}(q_{\text{current}}, q_{\text{next}}, \mathcal{C}_{\text{robot}})$\;
        $FirstTime \gets \text{False}$\;
    }{
        $q_{\text{SendServo}} \gets \text{ExecuteTrajectory}(\sigma(\Xi_{\text{current}}, \tau))$\;
        $q_i, T_i \gets $ \text{extra all points from buffer}\;
        $\dot{q}_{\text{current}}, \ddot{q}_{\text{current}} \gets$ calculate by $\sigma(\Xi_{\text{current}}, \tau)$\;
        $\sigma(\Xi_{\text{next}}, \tau) \gets$ \\
        \nonl $\text{MinStretchSpline}(q_i, \dot{q}_{\text{current}}, \ddot{q}_{\text{current}}, T_i, \mathcal{C}_{\text{robot}})$\;
        $\sigma(\Xi_{\text{current}}, \tau) \gets \sigma(\Xi_{\text{next}}, \tau)$\;
    }
}
$StartServo \gets \textbf{Monitor Thread}$\;
\nonl \textbf{Execute Trajectory} Divide the duration into $\Delta t_{\text{output}}$ intervals and compute the corresponding joint values for each.

\nonl \textbf{Monitor Thread:} Set $StartServo \gets \text{False}$ if teleoperation is stopped\;

\nonl \textbf{Sender Thread:} Send $q_{\text{SendServo}}$ to motor controller every $\Delta t_{\text{output}}$\;
\end{algorithm}

As shown in \text{Algorithm} \ref{precise}, when the system receives \n{inputs} from the Preprocess module, the StartServo flag is set to true, initiating the trajectory generation and execution (Line 1). The planning process proceeds in two phases\n{,} depending on the execution state. In the initial phase, the first target point is handled specially (Line 5-7), as it may differ significantly from \n{the} current position of \n{the} robot. The system uses the point-to-point method (SolvePTP, using the simplified closed-form solution for static transitions in \ref{tg})(Line 6), to smoothly guide the joints to the initial target, generating the trajectory $\sigma(\Xi, \tau)$. This trajectory is then executed(in the next loop), and $q_{\text{SendServo}}$ is calculated(Line 9). In subsequent iterations (Line 9-13), the system extracts joint angles and calculates time steps using timestamps from the buffer (Line 9). 

The \textbf{MinStretchSpline} (implementing the general optimization method in \eqref{min-prob}) generates trajectories by integrating current kinematic states (position/velocity/acceleration) with buffered target sequences,   while dynamically adapting time steps to satisfy joint limits extracted from URDF models. Only the first trajectory segment is executed to support incremental planning. Typically, \textbf{ExecuteTrajectory} and \textbf{MinStretchSpline} run asynchronously: while \textbf{ExecuteTrajectory} executes the current segment, \textbf{MinStretchSpline} calculates the next one, improving efficiency.

Throughout all the process, the monitoring thread \textbf{\n{Monitor Thread}} dynamically updates the StartServo flag, allowing the servoing process to be paused or resumed as needed. Concurrently, the \textbf{Sender Thread} transmits the generated joint commands $q_{\text{SendServo}}$ to the motor controllers at a stable frequency $\Delta t_{\text{output}}$, ensuring consistent, high-fidelity performance for precise manipulation tasks.
\vspace{-2mm}
\subsection{Framework for Rapid Motion Task}
Time delay is a major obstacle to maintaining the stability of teleoperation systems\cite{6909347}, as it diminishes operator transparency \cite{spongHistoricalPerspectiveControl2022}. In precise mode, delays arise because the robot needs to move from the starting position to the first target point, and must execute each commanded point sequentially. Initially, the robot may be positioned far from the first target point, and as it moves toward that target, new commands continue to arrive. 

\begin{algorithm}[tb]
\small
\caption{Rapid Mode Trajectory Planning}
\label{rapid}
\let\oldnl\nl
\newcommand{\nonl}{\renewcommand{\nl}{\let\nl\oldnl}}
\KwIn{$buffer$, $\mathcal{C}_{\text{robot}}$, $\Delta t_{\text{servo}}$, $\Delta t_{\text{output}}$}
\KwOut{$q_{\text{SendServo}}$}

\textbf{Initialization:} $StartServo \gets \text{True}$, $\sigma(\Xi, \tau) \gets \text{empty}$

\BlankLine
\While{$StartServo = \text{True} \text{ or buffer is not empty}$}{
    $q_{\text{current}} \gets \text{GetCurrentPositions}$\;
    \eIf{$\sigma(\Xi, \tau)$ \text{ is empty}}{
        $q_{\text{first}} \gets \text{pop buffer}$\;
        $\sigma(\Xi, \tau) \gets \text{SolvePTP}(q_{\text{current}}, q_{\text{first}}, \mathcal{C}_{\text{robot}})$\;
    }{
        $q_{\text{SendServo}} \gets \text{ExecuteTrajectory}(\sigma(\Xi_{\text{current}}, \tau), \Delta t_{\text{servo}})$\;
        $q_{\text{new}} \gets \text{pop buffer}$\;
        $q_{\text{current}}, \dot{q}_{\text{current}}, \ddot{q}_{\text{current}}, q_{\text{end}} \gets \sigma(\Xi_{\text{current}}, \tau)$\;
        $q_i \gets \text{PlanJointPath}(q_{\text{current}}, q_{\text{end}}, q_{\text{new}})$\;
        $\sigma(\Xi_{\text{next}}, \tau) \gets$ \\
        \nonl \text{MinStretchSpline}$(q_i, \dot{q}_{\text{current}}, \ddot{q}_{\text{current}}, T_i, \mathcal{C}_{\text{robot}})$\;
        $\sigma(\Xi_{\text{current}}, \tau) \gets \sigma(\Xi_{\text{next}}, \tau)$\;
    }
}
$StartServo \gets \textbf{Monitor Thread}$\;
\end{algorithm}

To mitigate this, the rapid mode always prioritizes the latest issued waypoint, disregarding previously commanded points that have not yet been reached. As illustrated in \text{Algorithm} \ref{rapid}, the trajectory from the starting position to the first waypoint is fully planned (Line 5-6), but only a small segment is executed for a fixed time, $\Delta t_{\text{servo}}$(Line 8). Then, trajectory continuously adjusts to the latest issued waypoint (Line 8-13). However, generating trajectories directly from the robot's current position to the latest waypoint can cause rapid acceleration changes, leading to jitter and reduced control smoothness. To address this, Algorithm \ref{rapid} generates each new trajectory using three key points: current position($q_{\text{current}}$), the end point of the previous trajectory($q_{\text{end}}$), and the newly received waypoint($q_{\text{new}}$) (Line 9-11). This approach ensures gradual acceleration changes, preventing abrupt movements while maintaining responsiveness and operational stability.
\section{EXPERIMENTS}
This section introduces the experiments we conduct to evaluate UTTG. Specifically, we applied the system to three \n{different} robot platforms, and compared the accuracy and trajectory smoothness with other algorithms to demonstrate the efficiency of our method.

\subsection{Qualitative Experiment}
To evaluate the performance of UTTG framework across different platforms, we conducted three experiments on the systems shown in Fig. \ref{experiment setups}:
\begin{figure}[t]
     \centering
     \includegraphics[width=0.9\linewidth]{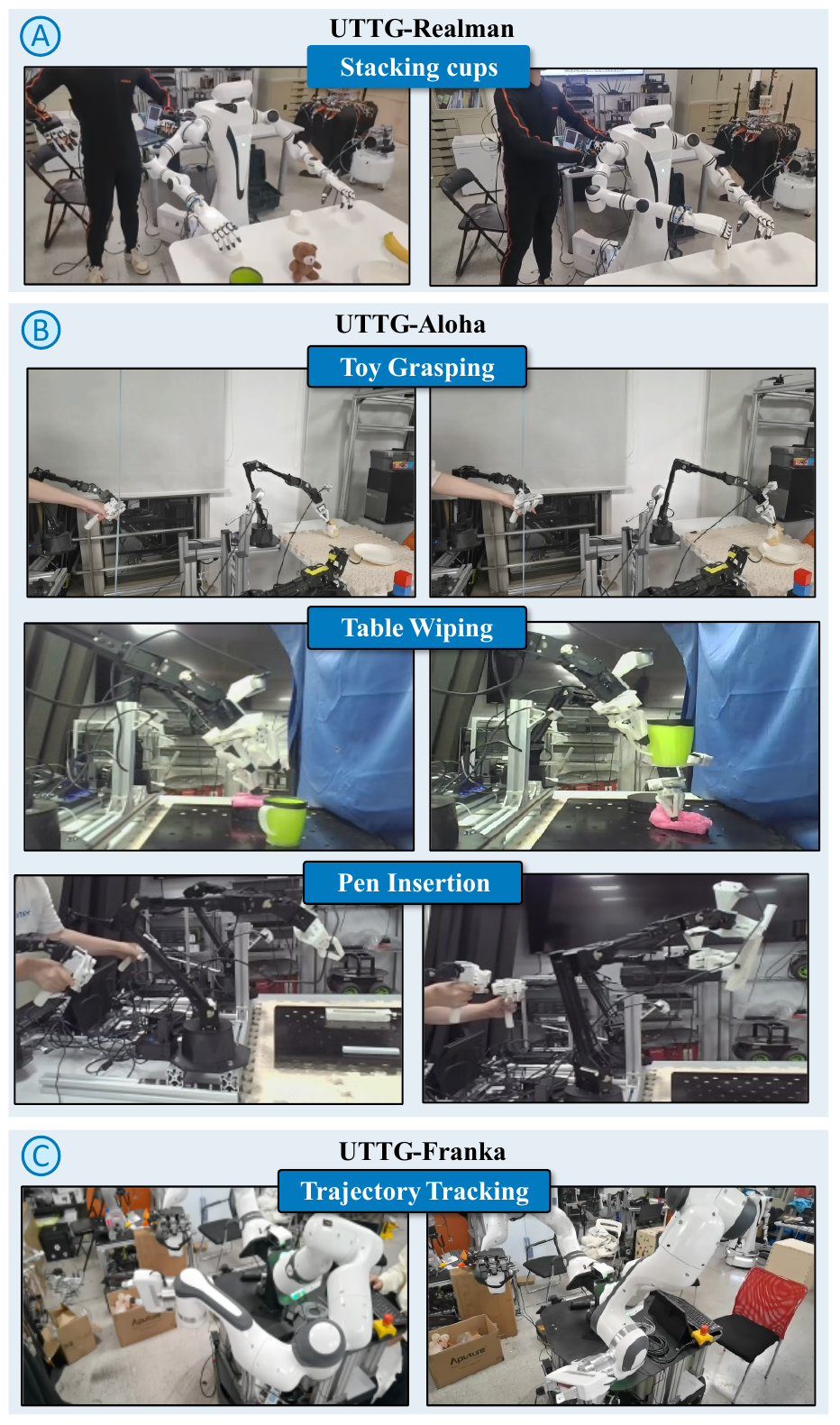}
     \caption{Snapshots in different platforms: (A) Realman with motion-capture suit, (B) Aloha using bimanual joint mapping, (C) Franka Research 3 Tasks. We refer readers to our video for better presentation. }
     \label{experiment setups}
     \vspace{-5mm}
 \end{figure}

\textbf{Trajectory Tracking:} Realman\footnote{Realman: \url{www.realman-robotics.cn}} executed predefined motions for trajectory accuracy evaluation.

\textbf{Precision Tasks:} Aloha performed three manipulation tasks:
\begin{itemize}
    \item \textit{Object Manipulation:} Soft toy and cylindrical can grasping with placement tasks.
    \item \textit{Pen Insertion:} Put the pen into the box.
    \item \textit{Table Wiping:} Dual-arm coordination: (i) cup displacement, (ii) table wiping, (iii) environment reset
\end{itemize}

\textbf{Dynamic Performance Evaluation:} The Franka platform was subjected to dynamic operational scenarios involving quick movement. 

Our framework exhibits dimensional independence via parallel joint-space processing, enabling complex multi-DoF applications. All experiments are conducted on an Intel NUC computer (i7-1165G7, 16GB RAM).
\begin{figure}[t]
    \centering
    \begin{subfigure}[b]{0.85\linewidth}
        \includegraphics[width=\linewidth]{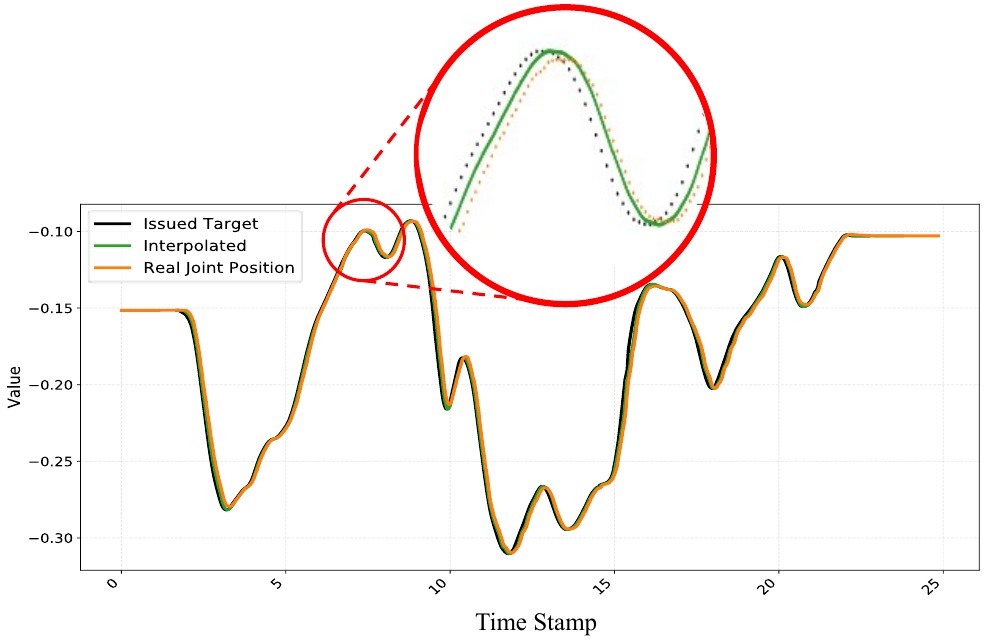} 
        \caption{UTTG framework trajectory}
        \label{ex1_uttg}
    \end{subfigure}
    \vspace{0cm}
    \begin{subfigure}[b]{0.85\linewidth}
        \includegraphics[width=\linewidth]{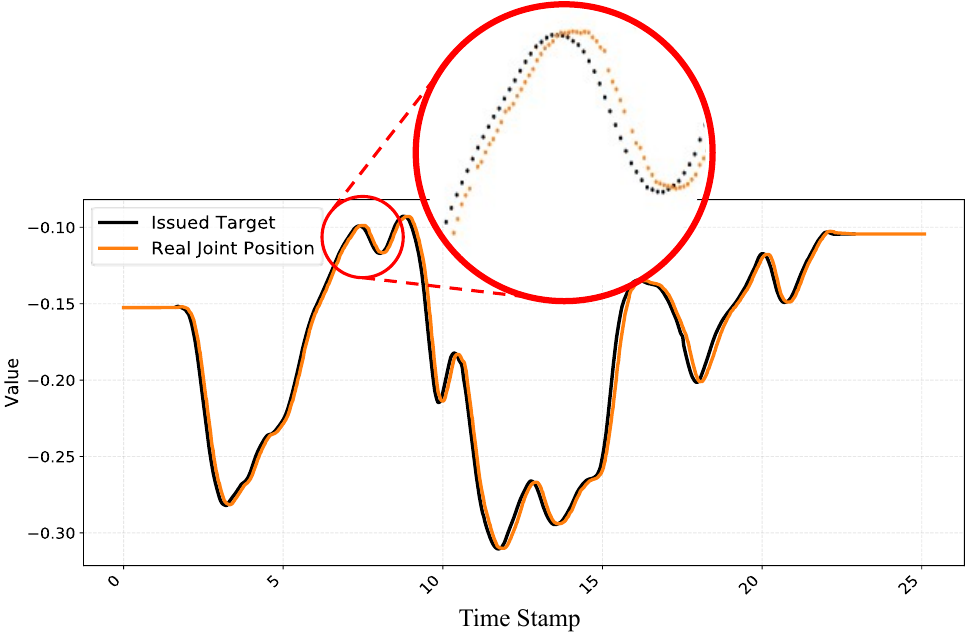} 
        \caption{Realman algorithm trajectory}
        \label{fig2b_ex1}
    \end{subfigure}
    \caption{\ref{draw_ex1}a and \ref{draw_ex1}b present the comparison of generated trajectories between UTTG framework and Realman algorithm. Our framework interpolates joint positions before sending commands to the motor, and the Realman algorithm optimizes and sends positions in the internal control loop, which is unreachable thus cannot be migrated to other robot platform.}
    \label{draw_ex1}
\vspace{-2mm}
\end{figure}
\vspace{-2mm}
\subsection{Comparison Result}
\subsubsection{Trajectory Tracking on Realman Robot}
This experiment evaluates tracking accuracy and trajectory smoothness. As shown in Fig. \ref{draw_ex1}, we compared the performance of our framework(\ref{ex1_uttg}) with the native control algorithm of Realman robot(\ref{fig2b_ex1}). Our framework interpolates the issued targets to high-ferquency joint trajectories, whereas the Realman algorithm processes raw low-frequency inputs. Analysis of the actuator feedback data (Real Robot Position) in magnified regions demonstrates that our framework generates trajectories with significantly smoother transitions and reduced abrupt changes, particularly during high-velocity segments.

\begin{table}
    \centering
    \caption{The task success rate with and without UTTG interpolation.}
    \begin{tabular}{ccccc}
    \toprule
     Case & Toy & Can & Pen & Table\\
    \midrule
    UTTG & 1 & 0.92 & 0.88 & 0.90\\
    No UTTG & 0.82 & 0.76 & 0.58 & 0.54\\
    \bottomrule
    \end{tabular}
    \label{Aloha}
    \vspace{-5mm}
\end{table}
\subsubsection{Precision Tasks on Aloha} 
We conducted 50 trials for each task under teleoperation with and without UTTG. The baseline method operated at 20Hz control frequency, while UTTG achieved 200Hz through online trajectory interpolation. To mitigate user-specific bias, two novice operators (each trained for 3 minutes) performed the tasks, with identical time constraints: each trial was limited to 10 seconds (Table wiping 15 seconds), and timeout was classified as failure. 

Table \ref{Aloha} illustrates the success rate across distinct tasks. For single-arm tasks (Toy/Can), both methods achieve comparable reliability due to inherent task simplicity. In complex scenarios requiring sustained dual-arm coordination (Table wiping/Pen insertion), baseline methods exhibit deteriorating reliability from jitter, while UTTG maintains robust performance through smooth trajectory generation.
\subsubsection{Dynamic Test on Franka}
\begin{table}
\centering
\caption{MAV Values Comparison (Units: rad/s\textsuperscript{2})}
\footnotesize
\setlength{\tabcolsep}{3.5pt}
\begin{tabular}{c|cc|cc|cc}
\hline
\textbf{Joint} & \multicolumn{2}{c|}{\textbf{No interpolation}} & \multicolumn{2}{c|}{\textbf{UTTG}} & \multicolumn{2}{c}{\textbf{Deoxys}} \\
\cline{2-7}
\textbf{No.} & Mean ($\omega$) & Std ($\alpha$) & Mean ($\omega$) & Std ($\alpha$) & Mean ($\omega$) & Std ($\alpha$)\\
\hline
0 & 6.930 & 1.252 & 0.473 & 0.323 & 1.441 & 0.952\\
1 & 8.585 & 1.275 & 0.623 & 0.325 & 0.717 & 0.413\\
2 & 9.003 & 1.359 & 0.776 & 0.291 & 1.794 & 0.432\\
3 & 4.288 & 1.224 & 0.370 & 0.293 & 0.554 & 0.265\\
4 & 33.585 & 3.008 & 1.841 & 0.573 & 2.887 & 1.114\\
5 & 4.672 & 1.267 & 0.288 & 0.269 & 0.300 & 0.203\\
6 & 34.434 & 3.083 & 1.869 & 0.568 & 2.545 & 1.107\\
\hline
\end{tabular}
\label{tab:mav_comparison}
\vspace{-3mm}
\end{table}
In this experiment, the capability to maintain motion quality is validated through joint acceleration analysis. Under the same trajectory, we computed the mean absolute value (MAV) of joint accelerations across different methods on the Franka platform. The acceleration MAV for joint $j$ is defined as: $\text{MAV}_j = \frac{1}{N}\sum_{i=1}^{N} |a_j(t_i)|$, where $a_j(t_i)$ denotes the $j$-th joint's acceleration at sample $t_i$. As shown in Table \ref{tab:mav_comparison}, our method demonstrates better performance advantages: the MAV values show substantial reductions compared to both non-interpolated baseline (92\%) and existing approaches (Deoxys control \cite{zhu2022viola}). The standard deviation metrics further indicate enhanced motion stability across all tested configurations.


\section{CONCLUSION}
This paper presents UTTG, a dimension-independent teleoperation framework that enables rapid cross-platform deployment and online continuous trajectory generation. \n{The purpose of this approach is to facilitate the potential release of low-cost devices for expert data collection, providing an effective tool for the construction of broader embodied intelligence datasets.} Experiments across three robotic platforms validate smoother trajectories (92\% MAV reduction) and better operation performance (36\% success rate increases in complex tasks) of our method.

For future work, we will integrate human motion prediction into our framework to \n{further} mitigate latency caused by hardware and communication delays. Additionally, we plan to develop reactive obstacle avoidance networks that cooperate with online trajectory generation, enabling real-time collision avoidance during teleoperation even when only end-effector inputs are available. 


\bibliographystyle{ieeetr}

\bibliography{ref}

\addtolength{\textheight}{-12cm}

\end{document}